\documentclass[12pt]{article}

\usepackage{times,color}
\usepackage{amsmath}
\usepackage{subfigure,epsfig,graphicx,epstopdf}
\usepackage[colorlinks]{hyperref}
\hypersetup{linkcolor = {black},	citecolor = {black},	urlcolor = {black}}
\usepackage{cite}

\newenvironment{sciabstract}{%
\begin{quote} \bf}
{\end{quote}}

\newcounter{lastnote}

\begin{document}

\title{Quantum Deep Learning for Mutant COVID-19\\ Strain Prediction}
\author{Yu-Xin Jin$^{1,2,3,4\ast}$,  Jun-Jie Hu$^{1,2,3\ast}$, Qi Li$^{5,6,\ast}$,\\ 
Zhi-Cheng Luo$^{1,7}$, Fang-Yan Zhang$^{1,5}$, Hao Tang$^{1,2,3\dagger}$,\\
Kun Qian$^{5\dagger}$, Xian-Min Jin$^{1,2,3\dagger}$ \\
\normalsize{$^1$TuringQ Co., Ltd., Shanghai 200240, China}\\
\normalsize{$^2$Center for Integrated Quantum Information Technologies (IQIT),}\\
\normalsize{School of Physics and Astronomy and State Key Laboratory of Advanced} \\
\normalsize{Optical Communication Systems and Networks,}\\
\normalsize{Shanghai Jiao Tong University, Shanghai 200240, China}\\
\normalsize{$^3$CAS Center for Excellence and Synergetic Innovation Center in Quantum}\\ \normalsize{Information and Quantum Physics,}\\
\normalsize{University of Science and Technology of China, Hefei, Anhui 230026, China}\\
\normalsize{$^4$ School of Mathematical Sciences, Nankai University, Tianjin, 300071, China}\\
\normalsize{$^5$Department of Information and Intelligence Development, Zhongshan Hospital,}\\
\normalsize{Fudan University, Shanghai 200032, China}\\
\normalsize{$^6$ Biomedical Engineering, Imperial College London,}\\
\normalsize{London, SW7 2BX, United Kingdom} \\
\normalsize{$^7$ Biological Science, Carnegie Mellon University, }\\
\normalsize{Pittsburgh, PA 15213, Pennsylvania, United State of America}\\
\normalsize{$^\ast$These authors contributed equally to this work}\\
\normalsize{$^\dagger$E-mail: htang2015@sjtu.edu.cn}\\
\normalsize{$^\dagger$E-mail: qian.kun@zs-hospital.sh.cn}\\
\normalsize{$^\dagger$E-mail: xianmin.jin@sjtu.edu.cn}\\
}

\date{}%\today

% Double-space the manuscript.
\baselineskip24pt

\maketitle

\begin{sciabstract}

New COVID-19 epidemic strains like Delta and Omicron with increased transmissibility and pathogenicity emerge and spread across the whole world rapidly while causing high mortality during the pandemic period. Early prediction of possible variants (especially spike protein) of COVID-19 epidemic strains based on available mutated SARS-CoV-2 RNA sequences may lead to early prevention and treatment. Here, combining the advantage of quantum and quantum-inspired algorithms with the wide application of deep learning, we propose a development tool named DeepQuantum, and use this software to realize the goal of predicting spike protein variation structure of COVID-19 epidemic strains. In addition, this hybrid quantum-classical model for the first time achieves quantum-inspired blur convolution similar to classical depthwise convolution and also successfully applies quantum progressive training with quantum circuits, both of which guarantee that our model is the quantum counterpart of the famous style-based GAN. The results state that the fidelities of random generating spike protein variation structure are always beyond 96$\%$ for Delta, 94$\%$ for Omicron. The training loss curve is more stable and converges better with multiple loss functions compared with the corresponding classical algorithm. At last, evidences that quantum-inspired algorithms promote the classical deep learning and hybrid models effectively predict the mutant strains are strong.\\

\end{sciabstract}

\section*{Introduction}\label{sec1}
Severe acute respiratory syndrome coronavirus 2 (SARS-CoV-2), the causal pathogen of coronavirus disease 2019 (COVID-19), has infected over 365 million people and caused over 5.65 million deaths worldwide as of January 28, 2022\cite{bib5}. Even though numerous vaccines have been proven efficient to suppress the severity and infectivity of the virus, multiple new SARS-CoV-2 variants emerge rapidly and put great therapeutic pressure. As reported, B.1.617.2 (Indian, Delta) strain is now the global dominant variant of concern, which shows stronger transmissibility and causes higher risk of hospitalization, intensive care and death\cite{bib6}. Moreover, B.1.617.2 strain weakens the antibody's ability to neutralize the virus and increases the possibility for convalescent patients to reinfect. In December 2020, the emergence of B.1.1.529 (South African, Omicron) dramatically boosts the spreading of the virus and was proven to cause a significant decrement on the efficacy of vaccines. Other variants of concern including B.1.1.7 (British, Alpha), B.1.351 (South African, Beta), and P.1 (Brazilian, Gamma) also create chaos for the global pandemic control. Therefore, there is an urgent need to predict the variation structure of COVID-19 epidemic strains like Delta and Omicron in order to take action of early prevention and quick treatment, thus contributing to the global pandemic control\cite{bib1,bib2,bib3,bib4}.\par

Quantum computation\cite{bib7} has attracted considerable interest in both academia and industry and been widely applied to the field of machine learning. Since near-term quantum devices are still fairly noisy, researchers mainly focus investigations on various possible hybrid quantum-classical machine learning algorithms\cite{bib15}. As an accelerator in heterogeneous computing architecture, photonic and quantum processors directly improve the performance of artificial neural network and play the role of domain specific processors. In order to bring the advantages of quantum processor into a wide range of deep learning applications, it is necessary to develop a high-level library and other compiler software. Our self-developed DeepQuantum aims to make it easy to develop quantum applications and get the help of artificial intelligence community. On that foundation, hybrid quantum-classical models including variational quantum eigensolvers, quantum convolutional neural networks\cite{bib8}, quantum generative adversarial networks(GAN)\cite{bib9,bib32} and quantum reinforcement learning\cite{bibQRL} have been developed. Additionally, quantum style-based generative adversarial networks will be proposed to better predict COVID-19 epidemic strains.\par

Style-based generative adversarial networks\cite{bib37,bib10,bib12} are successfully applied to generate high-resolution realistic images\cite{bib40} with training dataset style and enable intuitive, scale-specific control of the synthesis. We introduce a hybrid quantum-classical model associating scale-specific control ability with quantum inherent accelerating performance\cite{bib36} to optimize the training process of classic GAN. With the capability of feature control, style mixing is employed by switching from one latent code to another at the selected point in the synthesis network, thus generating a feature map equipped with mixing features.\par

To the greatest extent of maintaining quantum discriminator structure in our quantum style-based GAN, we also establish an operation called quantum-inspired blur convolution which is analogous to depthwise convolution in classical machine learning and it achieves quantum progressive training\cite{bib11}. Then, we turn the discriminator of style-based GAN into its quantum counterpart and adapt the original generator for SARS-CoV-2 RNAs. We compare classical style-based GAN and quantum style-based GAN by recording the training process with multiple loss functions\cite{bib13,bib14,bib31} and results show that the training of the quantum model is more stable, converges better as well as flexible to loss functions. Meanwhile, we use fidelity to evaluate the performance of predicting epidemic strains and the results are always over 96$\%$ for Delta, 94$\%$ for Omicron according to generated samples. Quantum style-based GAN, as an example, predicts the mutation probability of coronavirus strains by mining the past SARS-CoV-2 mutation RNA sequences, provides an effective way to realize the application of quantum-classical hybrid algorithms, and shows the advantages of quantum algorithms combined with deep learning.\par

\section*{Results and Discussion}\label{sec2}

\noindent\textbf{Datasets of mutated SARS-CoV-2 RNAs.} To collect various and essential SARS-CoV-2 mutation characteristics for quantum style-based GAN traning, we include 774 mutated SARS-CoV-2 complete genomes from 9 variants of concern consisting of B.1.1.7, B.1.351, B.1.429, B.1.525, B.1.526, B.1.617.1, P2, B.1.617.2 and B.1.1.529. Data come from publicly available databases GISAID\cite{bib17} and NCBI\cite{bib16}. We construct spike protein cohort by aligning complete genomes, intercepting spike protein fragment and squeezing duplicate sequences (described in “Methods”). In the hybrid quantum-classical model, we use 734 samples from the first 8 variants as the training cohort for generating and predicting Delta mutational sites, and add 40 Omicron samples to the previous cohort as the training cohort of Omicron mutational sites. The statistical result of the mutation frequency plot is split into four parts based on the spike protein position. \par

\hspace*{\fill}

\noindent\textbf{Quantum encoding.} To extract variation features of mutated SARS-CoV-2 RNAs and fit for the input of the hybrid quantum-classical model, we employ quantum encoding\cite{bib39} on the spike protein cohort. For each spike region gene sequence in the cohort, we compare it with the first identified SARS-CoV-2 numbered NC\_045512.2 in the corresponding position. If the above two RNA base sequences are different at a given locus, we mark 1 at the same position in a new vector and mark 0 if the opposite happens. Thus, we get a series of vectors containing number 0 and 1 in the size of 1024 (since we start training from 10 qubits).\par
Density matrix is an Hermite operator which needs to satisfy normalization, trace equal to 1 and semi-positive requirement. To meet these conditions, we first perform L2 normalization on the stated vector, then we compute its conjugate transposed vector and finally we multiply them in order to build a 10-qubit density matrix. For each spike region gene sequence we generate, similar quantum encoding is conducted(described in “Methods”).\par

\hspace*{\fill}

\noindent\textbf{The algorithm flow of quantum style-based GAN} We propose a hybrid quantum-classical model QuStyleGAN(quantum style-based GAN)\cite{bib34,bib38} for COVID-19 epidemic strain prediction, which directly takes spike protein cohort as the training dataset, to generate variation structure possessing SARS-CoV-2 mutation characteristics and then map it to a spike region gene sequence from one of Delta or Omicron strain (details described in”Methods”). The model comprises two key parts. One is the classical style-based generator and the other is the quantum progressive discriminator.\par
The classical style-based generator (Supplementary Fig.S1d) contains a mapping network consisting of 8 fully connected layers and a synthesis network consisting of 5 blocks ranging from resolution $4^{2}$ to $1024^{2}$ in steps of $4^{2}$, two convolution layers for each block. Detailed description is shown in Supplementary style-based GAN. The quantum progressive discriminator will be introduced later.\par
\begin{figure*}
%\centering
\includegraphics[width=1 \columnwidth]{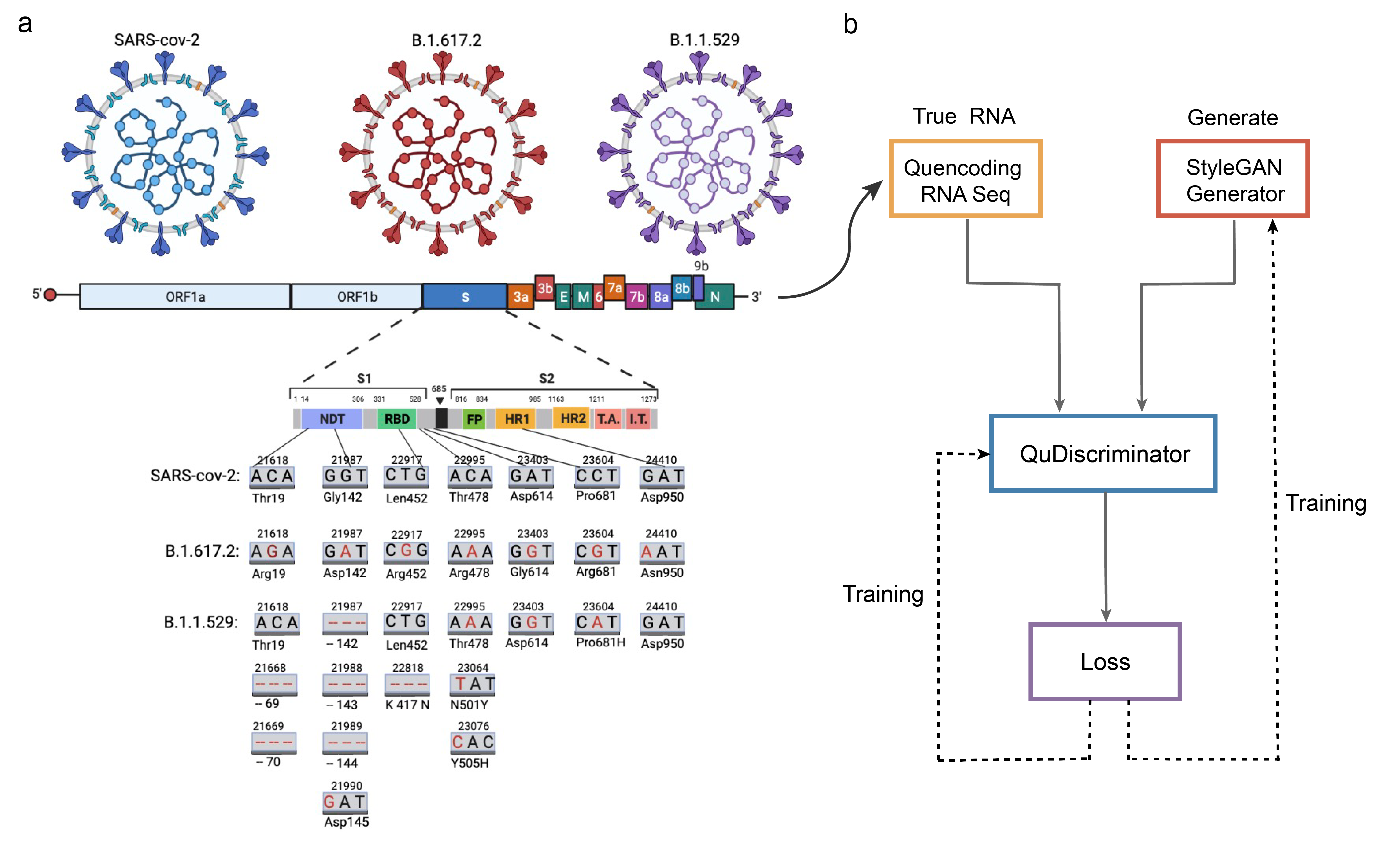}
	\caption{
	\textbf{Spike protein cohort and the overall architecture of hybrid quantum-classical model.} The length of spike protein from the cohort is 3822.
	\textbf{a}~The graphical presentation of phenotypic and genomic difference between SARS-CoV-2, B.1.617.2, and B.1.1.529 variant\cite{Biorender}. 
	\textbf{b}~The workflow of quantum style-based GAN one-step training.
	}
	\label{fig:1}
\end{figure*}

The workflow of quantum style-based GAN is shown in Fig.\ref{fig:1}. We first employ quantum encoding on the spike protein cohort to get true RNA mutation sequences. At the same time, the classical style-based generator generates fake RNA mutation sequences. Then, two kinds of data are fed in the quantum progressive discriminator to get their true or false judgment scores. According to the loss function we compute, we update parameters both in the classical generator and the quantum discriminator. Thus, one-step model training has completed.\par

\hspace*{\fill}

\begin{figure*}
	\includegraphics[width=\columnwidth]{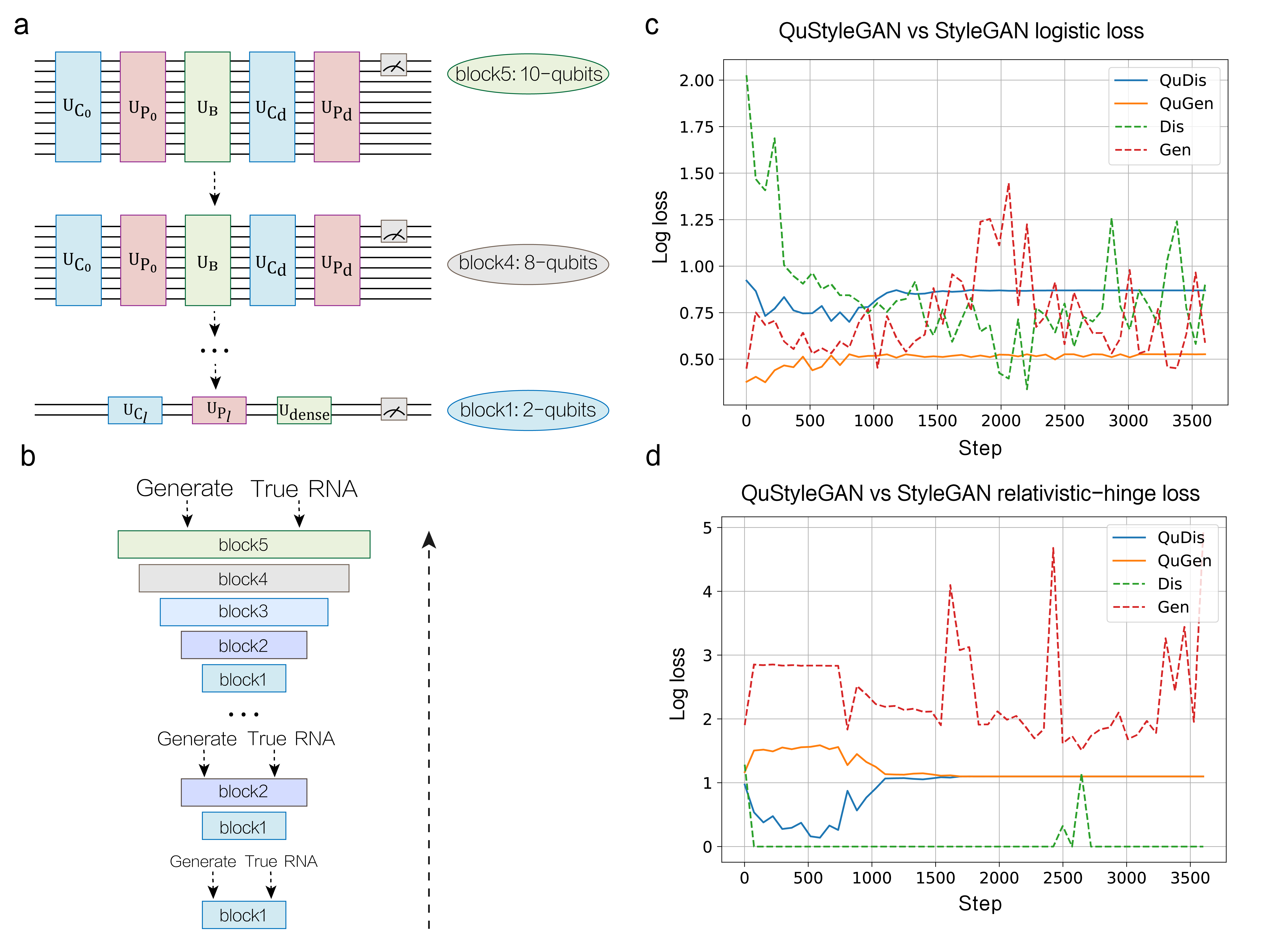}
	\caption{
	\textbf{Quantum progressive training and training process.}
	\textbf{a}~The quantum progressive discriminator network. Specifically, $U_{C_{0}}$ stands for the unitary matrix of the first quantum convolution layer for the first four modules while $U_{C_{d}}$ stands for the second quantum convolution layer associated with downscale; $U_{P_{0}}$ represents for the unitary matrix of the first quantum pooling layer while $U_{P_{d}}$ represents for the second quantum pooling layer; $U_{B}$ is defined by quantum-inspired blur convolution layer; $U_{C_{l}}$ and $U_{P_{l}}$ are the quantum convolution layer and the quantum pooling layer in the last module, respectively; $U_{dense}$ stands for the quantum dense layer.
	\textbf{b}~The quantum progressive training process. Specific blocks correspond to the quantum progressive discriminator.
	\textbf{c}~Log logistic loss for generator and discriminator of both quantum and classical style-based GAN.
	\textbf{d}~Log relativistic-hinge loss for generator and discriminator of both quantum and classical style-based GAN.
	}
	\label{fig:2}
\end{figure*}

\noindent\textbf{Quantum progressive discriminator and training.} The quantum progressive discriminator (Fig.\ref{fig:2}a, Supplementary Fig.S1 and QCNN) contains 5 quantum circuit modules ranging from 10 qubits to 2 qubits in steps of 2 qubits\cite{bib33,bib35}. For each of the first four modules, we construct two quantum convolution layers, two quantum pooling layers and one quantum-inspired blur convolution layer (shown in next section) with the corresponding number of qubits. Each quantum layer based on quantum gates corresponds to a unitary matrix. We put these unitary  matrices in order as displayed in Fig.\ref{fig:2}a. Consequent operation after aforementioned unitary operators is partial trace represented by 2 qubits observation symbols. Partial trace trashes the information of 2 qubits in the density matrix while saving others, which achieves the reduction of density matrix in steps of 2 qubits. The last module of the quantum progressive discriminator contains a quantum convolution layer, a quantum pooling layer, a quantum dense layer and a 1-qubit observation. The output of this module will be the final discriminative result.\par
Quantum style-based GAN training adopts progressive growing methodology same as classical style-based GAN algorithm(Supplementary Classical progressive training). By progressively introducing pre-configured quantum layers, we start training with the 2-qubit block, and then add larger 2$n$-qubit blocks to the networks as visualized in Fig.\ref{fig:2}b. In order to fade new quantum layers in smoothly, we follow residual block operation and downscale density matrices to match the current 2$n$-qubit block. Different from conducting average pooling to downscale in classical progressive training, we use partial trace to downscale between 2$n$-qubit density matrices in quantum progressive training. Benefits of quantum progressive training include simplifying the original generation problem of mapping latent codes to spike protein variation structures, stabilizing training process and reducing training time as suggested in ProGAN\cite{bib11}.\par
By contrasting quantum and classical style-based GAN with progressive training process recorded by two kinds of loss functions (details in "Methods" and Fig.S2) every 74 steps, we may discover some advantages of quantum neural networks. In total, Fig.\ref{fig:2}c and  Fig.\ref{fig:2}d illustrate that quantum style-based GAN(QuStyleGAN) training loss is more stable and converges better than classical style-based GAN(StyleGAN). For logistic loss, we can see that classical generator and discriminator loss alternatively oscillate and never reach convergence during the whole 3500 steps. In contrast, quantum generator and discriminator loss first intensively oscillate but converge after 1000 steps and sustain slight oscillation. Similar situation happens in relativistic-hinge loss, we can find more apparent convergence in QuStyleGAN after 1500 steps. These findings confirm the power of quantum neural networks and the hybrid quantum-classical model could potentially yield a quantum advantage.\par

\hspace*{\fill}

\noindent\textbf{Quantum-inspired blur convolution.} While designing the corresponding quantum discriminator, there is a blur layer benefit for noise reduction and feature extraction in its classical counterpart. After a careful study, the blur layer is actually a group convolution with the same number of groups, input channels and output channels, also called depthwise convolution (Supplementary Group convolution and depthwise convolution). Depthwise convolution can reduce many parameters, make effective use of hardware resources, and finally concatenate all feature maps. Based on these features and quantum properties, we propose a new quantum layer called quantum-inspired blur convolution.\par
\begin{figure*}
	\includegraphics[width=\columnwidth]{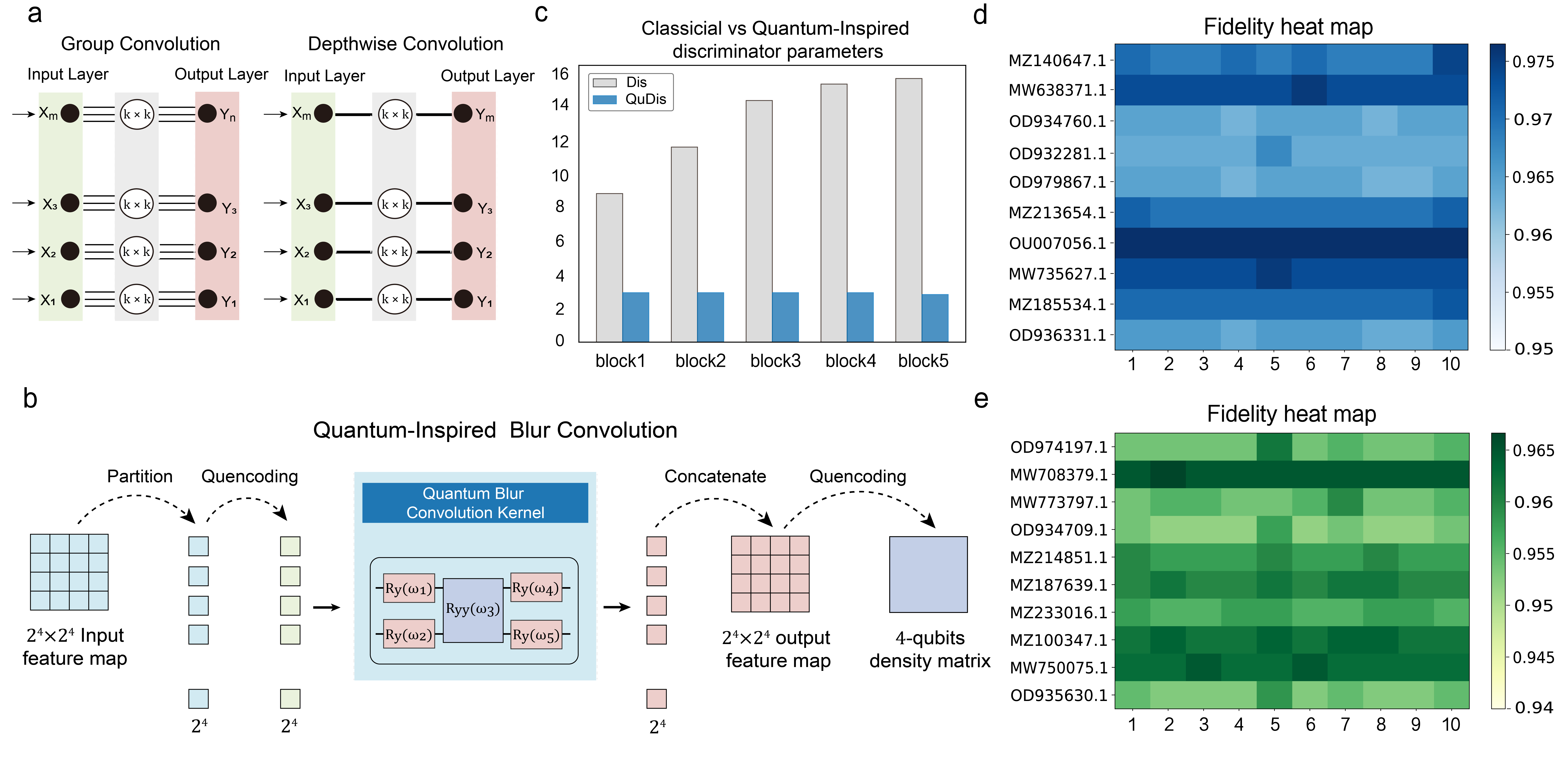}
	\caption{
	\textbf{Quantum-inspired blur convolution and performance evaluation.}
	\textbf{a}~Classical group and depthwise convolution.
	\textbf{b}~Quantum-inspired blur convolution including 3 main steps: partition, quantum blur convolution and concatenate operation.
    \textbf{c}~Log number of quantum and clssical style-based GAN discriminator parameters. Grey color bar represents for classical discriminator while blue color bar represents for the quantum discriminator.
    \textbf{d}~Heatmap consists of the fidelity between ten SP gene sequences in y-axis and ten generated Delta SP variation structure from x-axis. Color bar is associated with value of fidelity.
    \textbf{e}~Heatmap consists of the fidelity for generated Omicron SP variation structure.}
	\label{fig:3}
\end{figure*}

As shown in Fig.\ref{fig:3}b, conducting a quantum-inspired blur convolution needs three key steps, including partition, quantum blur convolution and concatenate operation. For simplicity, we take a feature map with the size of 4 qubits (any 2$n$ qubits can be applied in the same way) as example. For the partition step, we first divide the $2^{4} \times 2^{4}$ feature map into some $2^{2} \times 2^{2}$ feature maps; then we quantum encode these feature maps into some 2-qubit density matrices. For the quantum blur convolution step, we first construct a 2-qubit parameterized quantum circuits as the quantum blur convolution kernel; then we use the kernel to evolve with density matrices. For the last concatenate step, we put evolved density matrices into a $2^{4} \times 2^{4}$ feature map; then we quantum encode the feature map into a 4-qubit density matrix. Thus, a complete quantum-inspired blur convolution is done.\par

Since this quantum-inspired algorithm is an analog of depthwise convolution, it's possible to take a brief comparison between them. With the perspective of the whole process, as shown in the right part of Fig.\ref{fig:3}a, input tensor is divided into $m$ feature maps and then each feature map conducts convolution using a single $k \times k$ kernel; finally, all output feature maps are concatenated as the output tensor with the size $m$. Obviously, as stated above, this algorithm flow has three corresponding steps to depthwise convolution. From another point view of reducing parameters and saving resources, the left part of Fig.\ref{fig:3}a shows group convolution with $m \times k \times k \times \frac{n}{g}$ parameters where $g$ is the number of groups while depthwise convolution only needs $k \times k \times m$ parameters. Similarly, in this case as displayed in Fig.\ref{fig:3}b, the quantum-inspired module can be conducted in a 2-qubit parameterized quantum circuits with 5 parameters while normal quantum convolution has to be operated on a 4-qubit parameterized quantum circuits with more parameters. Therefore, as the expansion of quantum circuits consumes computing resources exponentially, quantum-inspired blur convolution in fact saves the computing resources and storage space greatly.\par

\hspace*{\fill}

\noindent\textbf{Quantum model evaluation.} The above quantum structure innovations are applied to the components of our quantum style-based GAN. First, we successfully solve quantum encoding of gene sequences. No matter what format the data is, vector or matrix, we both develop a method to encode the data to a density matrix with the corresponding size. This method may inspire further research. Second, quantum progressive training is perfectly carried out in our model. By training the quantum block-wise network progressively, we can save much training time and enable stable training process. It also suggests a possible way to train a quantum network with relatively large qubits. Third, we design a new quantum layer called quantum-inspired blur convolution. This quantum-inspired blur convolution layer reduces parameters, saves computing resources and serves as depthwise convolution in quantum machine learning expecting having the ability of noise reduction and feature extraction. This creative construction not only corresponds to classical style-based GAN, but also can be widely used like a normal quantum layer in later research.\par
Quantitatively, we evaluate our self-developed quantum GAN for variation structure prediction with spike protein cohort by fidelity. We randomly choose ten generated variation structures and ten gene sequences in spike protein cohort to compute the fidelity heatmap. Although it suggests the number of quantum discriminator parameters is far less than that in classical discriminator, the heatmap indicates that all fidelities of selected generated sequences are over 96$\%$ for Delta (Fig.\ref{fig:3}d), and 94$\%$ for Omicron (Fig.\ref{fig:3}e).\par
As shown in Fig.\ref{fig:3}d, the darkest blue line lies in the fidelities between ten random generated Delta spike protein sequences and OU007056.1 from strain P2, which means these generated sequences have the same highest fidelity among the heatmap and they are more similar to OU007056.1. The lightest blue spatial parts show the lowest fidelity in the heatmap, which mainly concentrate in the fidelities between the forth, eighth and ninth generated sequences and OD934760.1 from B.1.17 as well as OD979867.1 from B.1.525. However, high fidelity could be the result of little mutation compared with the original SARS-CoV-2 and the opposite situation may contribute to low fidelity. Additionally, we can make conclusions about horizontal lines, for instance, the sixth generated sequences is more similar to MW638371.1 from B.1.429 than others in the second line.\par
In the same way, Fig.\ref{fig:3}e suggests the highest fidelity appears in the cross point between the 2nd generated Omicron SP sequences and MW708379.1 from B.1.526 while the lowest fidelity exits in the forth line with OD934709.1 from B.1.17.\par

\hspace*{\fill}

\begin{figure*}
	\includegraphics[width=\columnwidth]{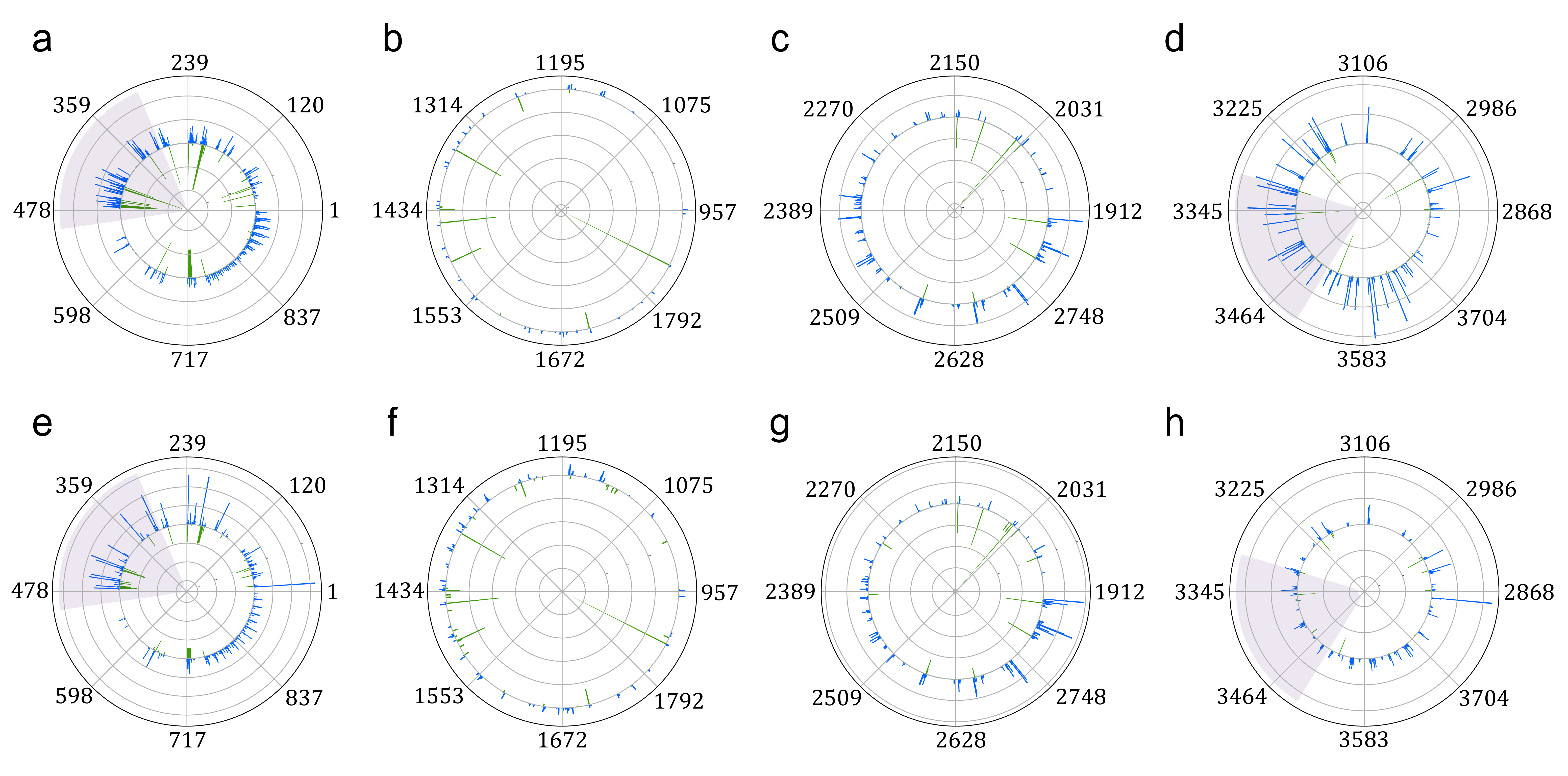}
	\caption{
	\textbf{Mutation frequency of spike protein cohort and generated spike protein.} The green color bar represents for spike protein cohort mutation frequency while the blue color bar represents for generated spike protein mutation frequency. The highlighted pink fan-shaped sector provides a clear indication on the range of the most frequent mutations in generated sequences.
	\textbf{a-d}~Mutation frequency of spike protein cohort and generated Delta spike protein, which in a 1-956 loci, in b 957-1911, in c 1912-2867, and in d 2868-3822.
	\textbf{e-h}~Mutation frequency of spike protein cohort and generated Omicron spike protein, which in the same loci subsections as Delta samples.}
	\label{fig:4}
\end{figure*}

\noindent\textbf{Performance evaluation of generated spike protein variation structure.} Apart from the fidelity between our self-developed model generated spike protein sequence and the selected variants spike protein sequence, a total of 1000 output sequences generated from the model are presented to show their specific mutation of positions in both the whole sequence and the spike protein sequence\cite{bib18}. From Fig.\ref{fig:1}a, Delta variant and Omicron variant coexist some single mutational sites but still follow its own distinctive pattern of mutation. Single nucleotide substitutions detected the already existing Delta and Omicron samples on the spike protein sequence region are accumulated around 300-500 as the first peak and around 2000 as the second peak. From Fig.\ref{fig:4}, the positions with high frequencies in the generated diagram emerges on several peaks at the range of 300-500 and 3300-3500. It has a relatively similar pattern at the first peak with the existing samples, but the newly generated sequences reveal a minor deviation from the second peak positional information of the mutation.\par

From the mutational frequency result of the existing Delta variant sequences, the peak of the position at 300-500 in the nucleotide sequence is located in the region of the S1 upstream to the RBD region\cite{bib19,bib20}, and the peak around 2000 in the nucleotide sequence is located at the RBD region of the spike protein\cite{bib1,bib21}. Within the RBD, the spike protein mutation on amino acid position 501 has the potential to increase ACE2-binding affinity by increasing the time of the “open conformation” of the ACE2 binding receptor. The mutation on amino acid position 519 is believed to decrease the convalescent serum neutralization in order to affect the efficiency of the vaccine and antibody detection in the body. The mutation on amino acid position 614 shows an increment on the viral load and produces more infectious viral individuals in human in vivo\cite{bib22,bib23}. Each of the mutations are proven with quantitative analysis on the ability to increase infectivity or avoidance of human immune system\cite{bib24,bib25,bib26}.\par

From the mutational frequency result of the Omicron of the input samples, Omicron contains more amino acid deletions and mutational sites on the spike protein than Delta\cite{bib41,bib42}. Omicron has the same mutational sites on 478 and 501, which gives this variant the ability to increase ACE2 binding affinity. The extra mutation on 484 has the potential to cause immune escape, and the deletion on 69 and 70 is responsible for the cause of S-gene target failure (SGTF)\cite{bib42}.\par

On the biological perspective to evalute the model's effectiveness, we use information of generated sequences from the model to develop an algorithm on R programming for integrating datas into a visual presentation of the mutational frequency for each potential site using bar plots. Based on the significant mutation on positions in the whole sequences and spike protein sequences, the difference or similarity between the generated and existing variants will be deducted and concluded to prove the possibility of the predicted variants and the variability of the model\cite{bib27,bib36}. As from the result frequency map, 300-500(Fig.\ref{fig:4}a and \ref{fig:4}e), 3300-3500(Fig.\ref{fig:4}d and \ref{fig:4}h) are considered as the most likely range with a single nuleotide variation.\par

Based on the result of generated sequences on Delta variant in Fig.\ref{fig:4}a, the most favorable sites of mutation occurs at the range of 300-500 among the generated sequences. Around this range of position in the nucleotide sequence, it is located in the region of the S1 upstream to the RBD region\cite{bib19,bib20}, which indicates that 300-500 has no clear structural function and does not participate the construction of the protein\cite{bib29,bib30}. The plausible reason is that this region of the nucleotide is on the upstream of the S1 region and might work as nucleotide that translates into regulatory proteins. \par

After adding 40 omicron sequences into the 734 variant samples and running the model again, even though the input frequency map occurs no obvious transformation, the result of the generated sequence provides a slight fluctuation of the mutational sites on the spike protein sequence. In Fig.\ref{fig:4}e, besides the frequent mutational occurrence between 300-500, sequence at the start of the spike protein to 300 has the inclination of getting mutated. For sequences between 0 to 300 in the S1 region, it refers to the N terminal domain, which its role is still not well marked\cite{bib43,bib44}. However, the ability of identifying specific carbohydrate portion in human cells during the early attachment makes it a suspicious factor of increasing ACE2 binding affinity in human cells\cite{bib45}. \par

As shown in Fig.\ref{fig:4}b and \ref{fig:4}f, there is no significant mutational sites appearing between 1000 and 2000, which is consistent with the actual situation. \par

Based on Fig.\ref{fig:4}c and \ref{fig:4}g, the generated sequences for Delta and Omicron both have highly frequent mutations on the second half of the circle, which is around 2400 - 2800. The corresponding protein sequence is in the section of FP, HR1, and HR2 in S2 subunit that is presented in Fig.\ref{fig:1}a. FP region contains a high content of glycine and alanine, which combines with HR1 and HR2 to allow a better anchoring and viral entry into the host cell after RBD binds to the cell.\par

On Fig.\ref{fig:4}d and \ref{fig:4}h, both Delta and Omicron generate new mutation sites between 3300 to the end of the spike protein sequence, which refers to the section of transmembrane domain and cytoplasmic tail domain. Both parts are believed to be highly conservative and play an important role throughout relative coronavirus\cite{bib46}. However, based of insufficient informations and limited cognition towards SARS-CoV-2, it is hardly to make an arbitrary conclusion on whether the mutation is going to decrease infectivity of the virus or deform its mechanism to cause more severe damage.\par

As mentioned above, positions with higher frequency (within 300-500 and 3300-3500) exhibited by the prediction model is a substantial indication of the future evolution on the COVID-19 variants as the pandemic continues. With the assistance of the modeling, we had accomplished a task by providing a considerable significance on predicting possible mutations from a novel perspective. This result may shed light upon research on future trends of the mutation and can be applied to other virus strains. Provided a complete understanding on the function of each single region in the spike protein, and a more comprehensive knowledge towards the protein folding and other biological form change, our model is compatible to such additional knowledge and hence will become more complete and vigorous.\par

In this work, we build a quantum style-based GAN for predicting spike protein variation structure of COVID-19 epidemic strains with mutated SARS-CoV-2 RNA sequences. The model to predict future COVID-19 variants with distinctive differences on mutational sites after adding Omicron samples helps convince the practicability and exclude the disturbance of randomness. The performance also shows great potential of quantum machine learning. Using metrics in quantum information, we get high fidelity as shown in Fig.\ref{fig:3}, which usually surpasses 96$\%$ for Delta, 94$\%$ for Omicron. It means our generated variation structures of COVID-19 epidemic strains equip with both quantum model and biological significance. Furthermore, we obtain a stable and convergent quantum GAN training process which is not common in classical model.\par 

There are also some limitations and future work for this hybrid quantum-classical model. First, more mutated SARS-CoV-2 RNAs may be better for model training and it’s accessible as existence of multiple variants. Second, more sophisticated quantum encoding can be completed in the future. Third, we can replace existing mapping method with inserting Delta or Omicron cohort in a 10-qubit module in the principle of style mixing. Finally, we might also modify model generator to fit for gene sequences directly by leveraging transformer.\par

Fortunately, construction of framework DeepQuantum accelerates the development of our model. This framework serves as quantum version of PyTorch, which combines quantum theory with deep learning tightly. With the help of DeepQuantum, researchers could exploit more hybrid quantum-classical models in various application scenarios and bring the advantage of quantum algorithm into deep learning application.\par

\section*{Conclusion}\label{sec3}
In summary, quantum style-based GAN can successfully predict variation structure of COVID-19 epidemic strains and is promising to be applied to other zoonotic viruses.  Based on the result, the model generates both mutational sites in a frequency table with and without the addition of omicron samples respectively and demonstrates a clear differentiation with their distinctive characteristics. This model also includes some quantum innovations such as quantum progressive training and quantum-inspired blur convolution, validates the power of quantum networks because of high fidelities and stable training process.\par
More sophisticated conclusions may lie in the well combination of quantum algorithms and classical artificial intelligence. Our proposed model stabilizes GAN training process, which provides an example of promoting artificial intelligence(AI) with quantum computation. Moreover, with the help of quantum computation, some problems can be solved while it is difficult for classical deep learning methods. Also, the elapsed time of quantum algorithms with the same methodology and effect as classical modules reduce greatly. In addition, the participation of AI in quantum computation naturally enriches solutions to problems in quantum fields.\par
At last, DeepQuantum works as a simplest tool combining the torch programming of deep learning and the qiskit-style programming of quantum computing. It could be an effective bridge to communicate quantum computation and artificial intelligence, thus contributing to the development and construction of hybrid quantum-classical models with the advantage of quantum algorithm. \par

\section*{Methods}\label{sec4}
\noindent\textbf{Strain selection and alignment.} Before picking up appropriated variant strains, a wild-type sequence called severe acute respiratory syndrome coronavirus 2 isolate Wuhan-Hu-1, which was first recorded in March 2020, was used as a reference sequence for analysis and alignment. NCBI has documented this sequence and named it as NC045512. The prevalence of the COVID 19 cases has been boosted in the past few months with the emergence of two new VOCs, Delta and Omicron. Those two variants are specifically chosen for generating the prediction of their spike protein mutation frequency due to their accelerating spreading and relatively high hazard worldwidely. Based on Fig.\ref{fig:1}a, both of the variants contain a considerable amount of mutational sites that are known to strengthen the virus. We used GSA and GISAID databases accessing from May 30 to June 29 in 2021 to derive 734 mutated SARS-CoV-2 sequences from 8 different variants\cite{bib16,bib17}. After the explosion of the Omicron variant, we derived 40 good-quality Omicron samples and added them into the training set as the second prediction model. In order to consider the accuracy and precision, collected sequences need to be ensured with their completeness and have multiple reviews. \par
CoVariant and CovidCG are used to provide integrated and pre-analyzed data in good classified format to help us simplify our own data processing. Covariant, supported by the database of GISAID from the website, is used as a source of detailed description and protein model presentation on each of the variant\cite{bib17}. Single nucleotide variants frequency on the nucleotide sequence and its corresponding amino acid are presented as a bar plot diagram by CovidCG website. Co-occurence of different mutational sites are shown in CovidCG to provide a thorough interactive relationship between different sites.\par
Statistical computing software is used to do data analysis and investigate the relationship or summary of our data within our intention. The frequency distribution of the mutational sites within the selected sequences are measured in a bar-plot using R programming with an algorithm to selectively pick positions with a high frequency by setting thresholds. The selected genome was aligned using clustalo with multi-alignment setting to obtain the relative probability of getting a mutation in a particular position for the construction of prediction model.\par

\hspace*{\fill}

\noindent\textbf{Interception and compression of spike protein gene sequences.} We construct spike protein cohort to train the proposed hybrid quantum-classical model. After alignment with the first identified SARS-CoV-2 numbered NC\_045512.2, we intercept the information of 21563-25384 loci in 774 mutated SARS-CoV-2 complete genomes as spike protein gene sequences.\par
It’s easy to compute that each spike protein gene sequence contains 3822 loci, which is inconsistent with the size of 10-qubits as input. Therefore, we compress some redundant information to get final spike protein gene sequences with the size of 1024 matching the input of quantum networks. We first fix all loci with mutation by comparing spike protein gene sequences with NC\_045512.2 one by one. Then, we reserve positions adjacent to mutation loci. Finally, 1024 loci are supplemental by randomly selecting other locations and we record all the chosen positions. So far, we get spike protein cohort of 1024 loci.\par

\hspace*{\fill}

\noindent\textbf{Quantum encoding for generated feature maps.} For spike protein gene sequences as training cohort, we simply operate quantum encoding on vectors. However, for generated feature maps produced by quantum style-based GAN generator, we have more steps to do. Firstly, we need to take the diagonal of each generated feature map. By taking a square root of the diagonal, we get a generated vector and it can be applied to quantum encoding as previously described. In brief, we conduct L2 normalization and then multiply conjugate transposed form in order.\par

\hspace*{\fill}

\noindent\textbf{Variation structure generating and mapping.} Variation structure information learned by quantum style-based GAN is stored in generated feature maps. In order to extract accurate and reliable mutation information as well as meet biological significance, we take the following action.\par
By investigating similarity between 774 mutated SARS-CoV-2 spike protein gene sequences and NC\_045512.2, we suppose 98\%(97\%) as the similarity of generated Delta(Omicron) spike protein gene sequences. In other words, in each generated spike protein gene sequence, only 21(31) loci mutate and other loci remain the same as NC\_045512.2. To obtain these mutation positions, we first get a probability vector by extracting square root of the diagonal of each generated feature map. Next, we sort values of this probability vector and choose top 21(31) indices as mutation positions. Yet, this is insufficient unless we return these mutation positions to original spike protein gene sequence with 3822 loci or even the complete genome. In fact, we have already done such jobs as we give mutation positions of the above three types(1024, 3822 and whole) for a generated variation structure.\par
Mapping generated variation structure to COVID-19 epidemic strains follows a principle similar to the above. Frankly, we randomly choose some aligned spike protein gene sequences of COVID-19 epidemic strain Delta or Omicron and then map generated variation structure onto them in the sense of mutation positions. In this way, we obtain some new mutated gene sequences possessed of variation characteristics of SARS-CoV-2 variants.\par

\hspace*{\fill}

\noindent\textbf{Torch-based quantum deep learning.} We implement our hybrid quantum-classical model QuStyleGAN by self-built torch-based software framework: DeepQuantum. DeepQuantum is established in the aim of combining quantum computing with machine learning and providing a powerful tool for constructing hybrid quantum-classical models in interdisciplinary efficiently. It is designed with following characteristics:
\begin{itemize}
    \item \textbf{QML application platform.} QML is the abbreviation of quantum machine learning. DeepQuantum combines quantum information with machine learning tightly because basic quantum structures are defined by torch variables like torch.tensor and some torch operators. Based on this natural superiority, DeepQuantum could be an ideal application platform for developing various hybrid quantum-classical models and exploring quantum advantages.
    \item \textbf{Simultaneous parameter optimization.} By leveraging torch calculation diagram, DeepQuantum can construct hybrid quantum-classical models with seamless backpropagation among classical and quantum neural networks. Meanwhile, parameterized hybrid quantum-classical models can be trained and update parameters via classical torch optimizer simultaneously.
    \item \textbf{Facilitate interdisciplinary.} Quantum machine learning is able to combine with various disciplines. DeepQuantum provides powerful tools to promote interdisciplinary such as prediction of binding energy by quantum mutual learning and prediction of gene interaction by quantum variational circuits.
\end{itemize}\par

\hspace*{\fill}

\noindent\textbf{Fidelity.} We evaluate our hybrid quantum-classical model by fidelity from quantum information. Gene sequences from spike protein cohort are notated as $\rho_{\text {in }}$ after quantum encoding while generated spike protein gene sequences are notated as $\rho_{\text {out }}$ after quantum encoding. We use the following formula\ref{eq:1} to calculate the fidelity
			\begin{equation}
			\label{eq:1}
			\operatorname{Tr}\left(\rho_{\text {in }} \rho_{\text {out }}\right)+\sqrt{\left(1-\operatorname{Tr}\left(\rho_{\text {in }}^{2}\right)\right)\left(1-\operatorname{Tr}\left(\rho_{\text {out }}^{2}\right)\right)}.
			\end{equation}
Also, we can use another form
			\begin{equation}
			\label{eq:2}
			\operatorname{Tr}\left(\sqrt{\sqrt{\rho_{i n}} \rho_{\text {out }} \sqrt{\rho_{\text {in }}}}\right).
			\end{equation} 
			
\hspace*{\fill}			
			
\noindent\textbf{Two kinds of loss functions.} We use logistic loss and relativistic-hinge loss to compare training process between quantum and classical style-based GAN. Here, we use formulas to elaborate on these two loss functions.\par
For quantum model, we use non-saturated logistic loss without gradient penalty as below:
			\begin{equation}
			\label{eq:3}
			\operatorname{Loss}_{G}=\log \left(e^{-D(G(z))}+1\right),
			\end{equation}
			\begin{equation}
			\label{eq:4}
			\operatorname{Loss}_{D}=\log \left(e^{D(G(z))}+1\right)+\log \left(e^{-D(x)}+1\right).
			\end{equation}
$G$ represents for generator, $D$ represents for discriminator, $z$ is the latent code and $x$ is the true data. For classical model, we add a penalty term to discriminator loss just for successfully training. The penalty term shows as below:
			\begin{equation}
			\label{eq:5}
			\frac{1}{2} \gamma_{1 \sum \nabla_{\text {real }}^{2}}+\frac{1}{2} \gamma_{2 \sum \nabla_{\text {faks }}^{2}}.
			\end{equation}
In experiment, we take $\gamma_{1}=10$ and $\gamma_{2}=0$.\par
For relativistic-hinge loss, both quantum and classical style-based GAN adopt the same form. We first compute two differences as below:
			\begin{equation}
			\label{eq:6}
			\operatorname{diff}_{r f}=D(x)-E(D(G(z))),
			\end{equation}
			\begin{equation}
			\label{eq:7}
			\operatorname{diff}_{f r}=D(G(z))-E(D(x)).
			\end{equation}
Then, we construct generator loss and discriminator loss:
			\begin{equation}
			\label{eq:86}
			\operatorname{Loss}_{G}=E\left(\operatorname{ReLU}\left(1+\operatorname{diff}_{r f}\right)\right)+E\left(\operatorname{ReLU}\left(1-\operatorname{diff}_{f r}\right)\right),
			\end{equation}
			\begin{equation}
			\label{eq:9}
			\text { Loss }_{D}=E\left(\operatorname{ReLU}\left(1-\operatorname{diff}_{r f}\right)\right)+E\left(\operatorname{ReLU}\left(1+\operatorname{diff}_{f r}\right)\right),
			\end{equation}
			\begin{equation}
			\label{eq:10}
			\operatorname{ReLU}(\mathrm{x})=\left\{\begin{array}{l}
            \mathrm{x}, \mathrm{x} \geq 0 \\
            0, \mathrm{x}<0.
            \end{array}\right.
			\end{equation}
$E$ represents for the mean value.

%Bibliography

\subsection*{Data availability}
The data that support the findings of this study are available from the corresponding authors on reasonable request.

\subsection*{Acknowledgments}
This research was supported by National Key R\&D Program of China (2019YFA0706302, 2019YFA0308700, 2017YFA0303700), National Natural Science Foundation of China (61734005, 11761141014, 11690033, 11904229), Science and Technology Commission of Shanghai Municipality (STCSM) (21ZR1432800, 20JC1416300, 2019SHZDZX01), and Shanghai Municipal Education Commission (SMEC) (2017-01-07-00-02- E00049). X.-M.J. and K.Q. acknowledge additional support from a Shanghai talent program.

%\subsection*{Author contributions} 
%X.-M.J. conceived and supervised the project. X.-M.J. performed the simulations and fabricated the photonic chips. X.-M.J. performed the experiment and analyzed the data. X.-M.J. interpreted the data and wrote the paper, with input from all the other authors.

\subsection*{Competing interests}
The authors declare no competing interests.

\baselineskip21pt

%============================fig1===========================
\par
%=====================================================

%============================fig2===========================
\par
%=====================================================

%============================fig3===========================
\par
%=====================================================

%============================fig4==========================
\par
%=====================================================

\clearpage

\section*{Supplementary Materials: Quantum Deep Learning for Mutant COVID-19 Strain Prediction}
\setcounter{table}{0}
\setcounter{equation}{0}
\setcounter{figure}{0}

\renewcommand{\thetable}{{S}\arabic{table}}
\renewcommand{\theequation}{{S}\arabic{equation}}
\renewcommand{\thefigure}{{S}\arabic{figure}}

\noindent\textbf{StyleGAN.} Style-based GAN(StyleGAN) is considered as a new generative adversarial network that was first proposed by NVIDIA after progressive GAN(ProGAN). Before that, ProGAN is used to solve the quality issues of generating images with high resolution, and its core ideology behind the technique is progressive training. It starts with a low-resolution 4x4 image generator and discriminator and gradually elevates the resolution of the image to develop the model to be able to construct more details and generate images with much higher resolution. However, the limitation of ProGAN is that it does not have the ability to control the specific features of the generated images. Therefore, if the random parameter that is inputted in the model has a slight modification, it will lead to a series of changes or deviations towards the final generated image. \par
StyleGAN comes into and perfectly solves this problem as it uses different layers to specify one of the random parameter in order to only manipulate a particular feature. Taking a face of human as an example, a rough scaled input can control features like gesture, facial shape, hair style, and glasses; A medium scaled input will be able to control some fine features like details on the hair and the closure of lips; When moving to the high scaled features, it is possible to even accurately control the color of the hair, the color of the skin, and even deblurring of the background to make the image more realistic.\par
The innovation of StyleGAN focuses more on the generator part. Instead of using latent code in traditional GAN, StyleGAN uses the output from mapping networks and adaptive instance normalization(AdaIN) to replace the bored and repetitive traditional inputs. Besides, it is believed that replacing the self-learned constant with the traditional inputs can improve the entanglement between features and increase the quality of the image.\par

\hspace*{\fill}	

\begin{figure*}
	\includegraphics[width=\columnwidth]{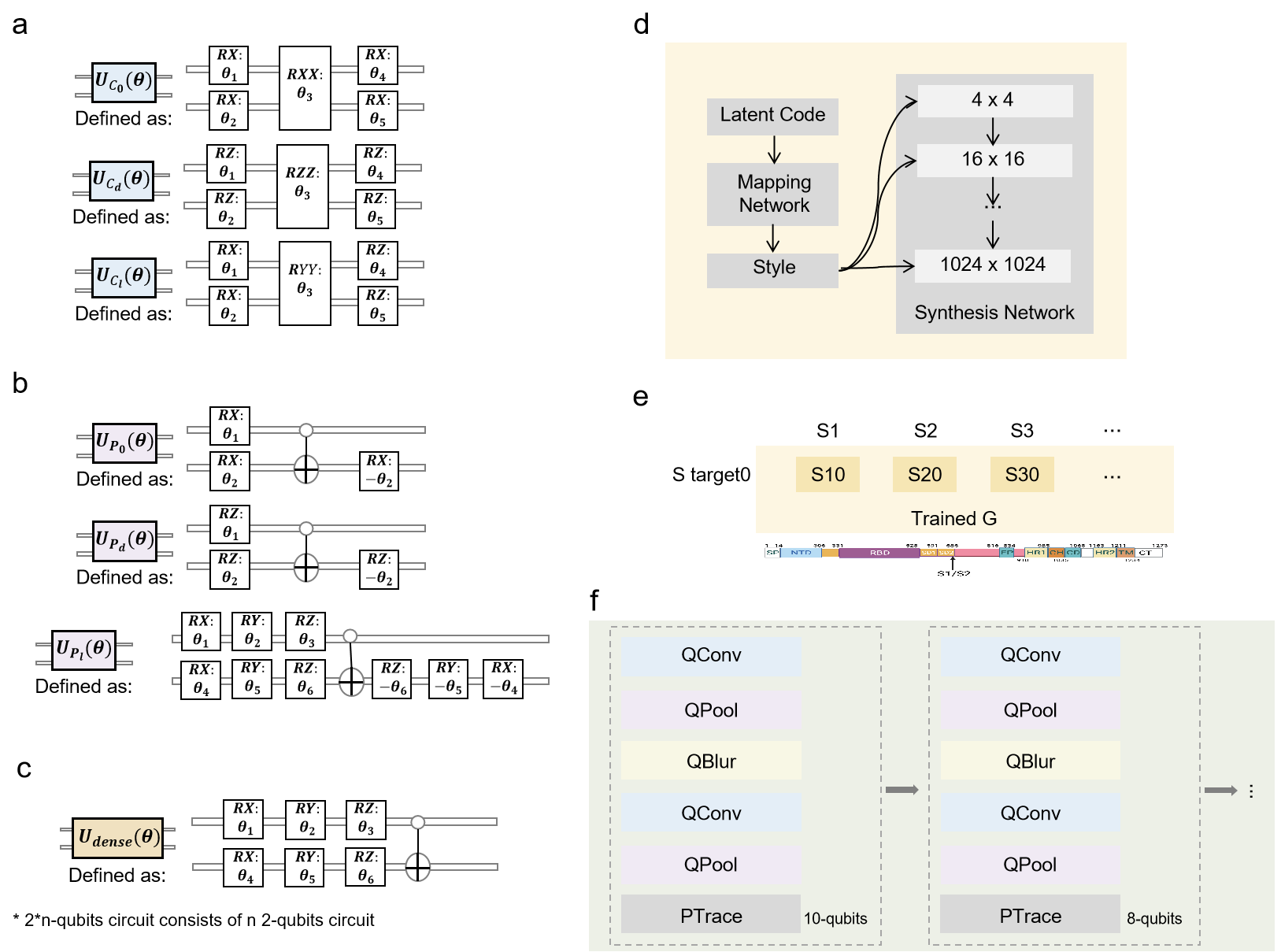}
	\caption{
	\textbf{Detailed quantum style-based GAN.}
	\textbf{a}~2-qubit QuConv.
	\textbf{b}~2-qubit QuPool.
	\textbf{c}~2-qubit Qudense.	
	\textbf{d}~StyleGAN Generator.
	\textbf{e}~Style Mixing RNA.
	\textbf{f}~Quantum Progressive Discriminator.
	}
	\label{fig:7}
\end{figure*}\par

\noindent\textbf{QCNN based on quantum circuits.} Quantum convolution networks mainly include quantum convolution layers, quantum pooling layers and quantum dense layers and consist of rotated x gate, rotated y gate, rotated z gate and CNOT gate. As shown in Fig.\ref{fig:7}, our quantum progressive discriminator uses three kinds of QuConv, three kinds of QuPool and one QuDense.\par
For the first four blocks, we construct them with the first quantum convolution layer defined as $U_{C_{0}}$, the first quantum pooling layer defined as $U_{P_{0}}$, quantum-inspired blur convolution layer, the second quantum convolution layer defined as $U_{C_{d}}$, the second quantum pooling layer defined as $U_{C_{d}}$ and a partial trace operation. For the last block, we use the last quantum convolution layer defined as $U_{C_{l}}$, the last quantum pooling layer defined as $U_{P_{l}}$ and the quantum dense layer defined as $U_{dense}$.\par

\hspace*{\fill}	

\noindent\textbf{Group convolution and depthwise convolution.} A grouped convolution, a set of convolutions, was first used to distribute over multiple GPUs to help a network learn multiple level features as an engineering compromise but found helpful in improving classification accuracy by ResNet and other models. Moreover, by exposing a new dimension to increase grouped convolutions and the cardinality, which refers to size of the set of transformations, it was examined applicable to increase accuracy.\par
In depth-wise convolution, each filter channel is only at one input channel. In the following example, 3 channel filters are presented. The depth-wise convolution is applied here to break the filter and image into three different channels and then coil and roll the corresponding image with corresponding channel. They then will stack on each other back. In order to produce same effect with normal convolution, depth-wise convolution is used to select a channel, make all the elements zero in the filter except that channel and then convolve. Three different filters are needed for each channel. Although parameters are remaining the same, this convolution will give out three output channels with only one filter as it is converged from three filters into one.\par

\hspace*{\fill}	

\begin{figure*}
	\includegraphics[width=\columnwidth]{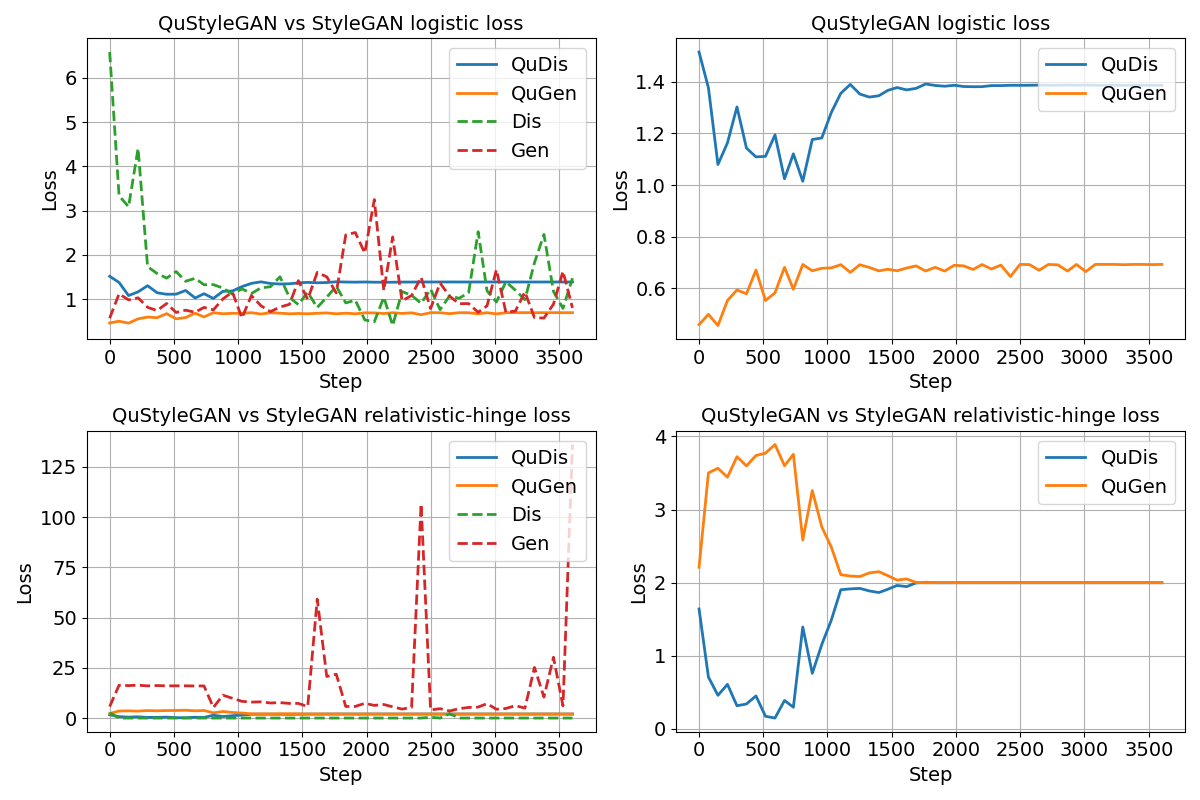}
	\caption{
	\textbf{Original training process of quantum and classical style-based GAN using logistic loss and relativistic-hinge loss.}
	}
	\label{fig:6}
\end{figure*}\par

\noindent\textbf{Classical progressive training.} Progressive GAN is an algorithm that StyleGAN generator and discriminator models are used to train their dataset. It can be interpreted as a way of gradually increasing the complexity of the input for the model to fit into. For example, a model dealing images with progressive GAN can start with small images like a 4x4 image. Until the model is fit and stable, both the discriminator and the generator of the model will expanded to twice of the size from its original size, like 8x8. After that, a new block will help building onto it to support this larger image size or dataset size. After a series of stabilization on the model, the model then again will train by expanding its dimension on the generator and discriminator to take over and deal with more complicated images, like 1024x1024.\par

\hspace*{\fill}	

% \begin{figure*}
% 	\includegraphics[width=1.8\columnwidth]{ps2.png}
% 	\caption{
% 	\textbf{Original training process of QuStyleGAN and StyleGAN using logistic loss and relativistic-hinge loss.}
% 	}
% 	\label{fig:6}
% \end{figure*}\par

\noindent\textbf{Original training process of quantum and classical style-based GAN.} In order to present the contrast of training process between quantum and classical models clearly, we made a logarithmic transformation of original loss in the article. We now display the original training process in Fig.\ref{fig:6}.

\end{document}